\documentclass[10pt,twocolumn,letterpaper]{article}

\usepackage{cvpr}
\usepackage{times}
\usepackage{epsfig}
\usepackage{graphicx}
\usepackage{amsmath}
\usepackage{amssymb}
\usepackage{amsthm}
\usepackage{multirow} 
\usepackage{diagbox} 
\usepackage[noend]{algpseudocode}
\usepackage{algorithmicx,algorithm}
\usepackage{multicol}
\usepackage{float}
\usepackage{caption}
\usepackage{indentfirst}

\usepackage[pagebackref=true,breaklinks=true,letterpaper=true,colorlinks,bookmarks=false]{hyperref}

\cvprfinalcopy 

\def\cvprPaperID{7098} 

\begin{document}

\title{Reusing Discriminators for Encoding:\\
Towards Unsupervised Image-to-Image Translation}

\author{
 Runfa Chen, Wenbing Huang, Binghui Huang, Fuchun Sun\thanks{Corresponding author: Fuchun Sun.} , Bin Fang\\
 Institute for Artificial Intelligence, Tsinghua University (THUAI)\\
 Beijing National Research Center for Information Science and Technology (BNRist),\\
 State Key Lab on Intelligent Technology and Systems,\\ Department of Computer Science and Technology, Tsinghua University, Beijing, P.R.China\\ 
 {\tt\small crf18@mails.tsinghua.edu.cn}, {\tt\small hwenbing@126.com}\\
  {\tt\small \{hbh18@mails., fcsun@, fangbin@mail.\}tsinghua.edu.cn}
 }

\maketitle

\begin{abstract}
   Unsupervised image-to-image translation is a central task in computer vision. Current translation frameworks will abandon the discriminator once the training process is completed. This paper contends a novel role of the discriminator by reusing it for encoding the images of the target domain. The proposed architecture, termed as NICE-GAN, exhibits two advantageous patterns over previous approaches: First, it is more compact since no independent encoding component is required; Second, this plug-in encoder is directly trained by the adversary loss, making it more informative and trained more effectively if a multi-scale discriminator is applied. The main issue in NICE-GAN is the coupling of translation with discrimination along the encoder, which could incur training inconsistency when we play the min-max game via GAN. To tackle this issue, we develop a decoupled training strategy by which the encoder is only trained when maximizing the adversary loss while keeping frozen otherwise.
   Extensive experiments on four popular benchmarks demonstrate the superior performance of NICE-GAN over state-of-the-art methods in terms of FID, KID, and also human preference. Comprehensive ablation studies are also carried out to isolate the validity of each proposed component. Our codes are available at \href{https://github.com/alpc91/NICE-GAN-pytorch}{https://github.com/alpc91/NICE-GAN-pytorch}.
\end{abstract}


\section{Introduction}
\vskip -5pt
Image-to-Image translation transforming images from one domain to the other has boosted a variety of applications in vision tasks, from colorization~\cite{zhang2016colorful}, image editing~\cite{dekel2018sparse}, super-resolution~\cite{ledig2017photo} to video generation~\cite{wang2018video}. Given the extensive effort of collecting paired images between domains, a more practical line of research~\cite{CycleGAN2017,liu2017unsupervised,huang2018munit,DRIT,Kim2020U-GAT-IT} 
directs the goal to unsupervised scenario where no paired information is characterized. 
Due to the non-identifiability problem~\cite{liu2017unsupervised} in unsupervised translation, various methods have been proposed to address this issue by using additional regulations including weight-coupling~\cite{liu2017unsupervised}, cycle-consistency~\cite{CycleGAN2017,kim2017learning,yi2017dualgan}, forcing the generator to the identity function~\cite{taigman2016unsupervised}, or more commonly, combination of them.


When we revisit current successful translation frameworks (such as the one proposed by CycleGAN~\cite{CycleGAN2017}), most of them consist of three components for each domain: an encoder to embed the input image to a low-dimension hidden space, a generator to translate hidden vectors to images of the other domain, and a discriminator for domain alignment by using GAN training~\cite{goodfellow2014generative}. While this piled-up way is standard, we are still interested in asking: \emph{is there any possibility to rethink the role of each component in current translation frameworks?} and more importantly, \emph{can we change the current formulation (for example, to a more compact architecture) based on our rethinking?}

\begin{figure}[t]
\setlength{\abovecaptionskip}{-7pt} 
\setlength{\belowcaptionskip}{5pt} 
\begin{center}
\includegraphics[width=1.0\linewidth]{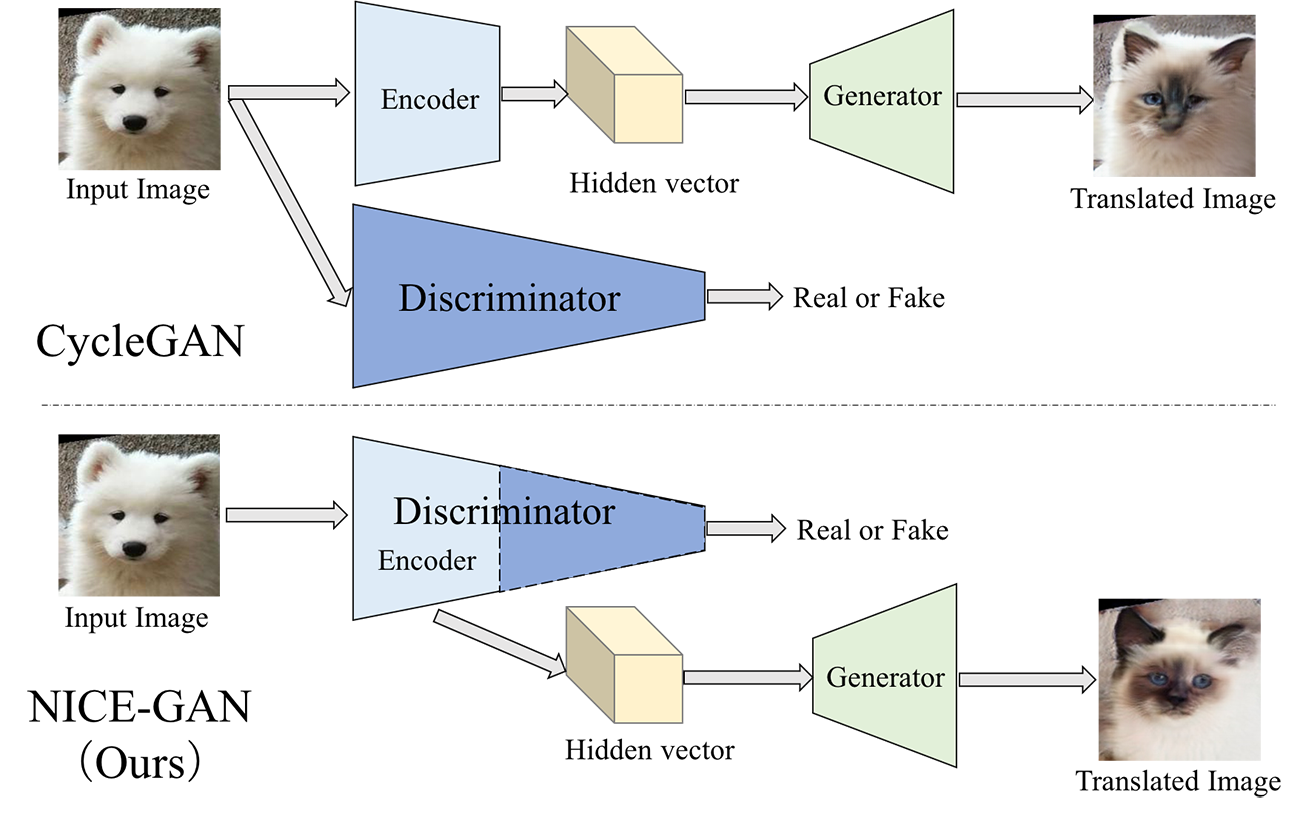}
\end{center}
   \caption{Illustrative difference between CycleGAN-alike methods and our NICE-GAN.}
\label{fig:title}
\end{figure}

The answer is yes, if we check the relation between the encoder and the discriminator. Basically, the discriminator is to distinguish between the translated image of the source domain and the real image of the target domain. To do so, the discriminator should conduct sort of semantics encoding of the input images before it can tell what images are true and what are false. This, in other words, contends the two roles of the discriminator: encoding and classifying. Indeed, the DCGAN paper~\cite{radford2015unsupervised} has revealed the encoding ability of the discriminator: strongly responses to the input image are observed in the first 6 learned convolutional features from the last convolution layer in the discriminator.

Upon the motivation mentioned above, this paper proposes to reuse the discriminator for encoding. In particular, we reuse early layers of certain number in the discriminator as the encoder of the target domain, as illustrated in Figure.~\ref{fig:title}. Such kind of reusing exhibits two-fold advantages: 
\textbf{I.} A more compact architecture is achieved. Since the encoder now becomes part of the discriminator, we no longer require an independent component for encoding. Also, unlike existing methods where the discriminator is abandoned after training, its encoding part is still kept for inference in our framework. 
\textbf{II.} The encoder is trained more effectively. Traditional training of the encoder is conducted by back-propagating the gradients from the generator, which is indirect. Here, by plugging it into the discriminator, the encoder is directly trained through the discriminative loss. Moreover, modern discriminators have resorted to the multi-scale scheme for more expressive power~\cite{durugkar2016generative,iizuka2017globally,demir2018patch,wang2018high}; our encoder will inherit the expressive ability by nature if the multi-scale discriminator is applied.

A remaining issue of our approach is how to perform adversary training. For traditional methods~\cite{CycleGAN2017,liu2017unsupervised,huang2018munit,DRIT,Kim2020U-GAT-IT}, the encoder is trained along with the generator for minimizing the GAN loss, while the discriminator is trained separately to maximize the objective. In our framework, the encoder and the discriminator become overlap, and it will bring in instability if we apply traditional training setting---the encoder as part of translation is trained for minimizing, and at the same time it belongs to the discriminator and is also trained for maximizing. To eliminate the inconsistency, we develop a decoupled training paradigm. Specifically, the training of the encoder is only associated with the discriminator, independent to the generator. Our experiments on several benchmarks show that such simple decoupling promotes the training remarkably (see details in Section~\ref{sec:training_exp}). 
Another intuition behind is that disentangling the encoder from the training of translation will make it towards more general purpose of encoding other than translation along, thereby enabling more generality.  

We summarize our contributions as follow.

\setlength{\parskip}{-6pt}
\begin{itemize}
\setlength{\itemsep}{2pt}
\setlength{\parsep}{0pt}
\setlength{\parskip}{0pt}
    \item To the best of our knowledge, we are the first to reuse discriminators for encoding specifically for unsupervised image-to-image translation. By such a reusing, a more compact and more effective architecture is derived, which is dubbed as No-Independent-Component-for-Encoding GAN (NICE-GAN).
    \item Given that the reusing of discriminator will incur instability in terms of typical training procedure, this paper develops a decoupled training paradigm, which is simple yet efficient.
    \item Extensive experimental evaluations on several popular benchmarks reveal that the proposed method outperforms various state-of-the-art counterparts. The comprehensive ablation studies are also conducted to verify the effectiveness of each proposed component.   
\end{itemize}
\setlength{\parskip}{0pt}

\section{Related Work}
\vskip -3pt

\noindent
\textbf{Image-to-image translation.}
Conditional GAN-based standard framework, proposed by Isola \etal \cite{isola2017image} , promotes the study on image-to-image translation. Several works extend it to deal with super-resolution\cite{wang2018high} or video generation\cite{wang2018video}. Despite of the promising results they attain, all these approaches need paired data for training, which limits their practical usage.

\noindent
\textbf{Unsupervised image-to-image translation.}
In terms of unsupervised image-to-image translation with unpaired training data, CycleGAN~\cite{CycleGAN2017}, DiscoGAN~\cite{kim2017learning}, DualGAN~\cite{yi2017dualgan} preserve key attributes between the input and the translated image by using a cycle-consistency loss. Various studies have been proposed towards extension of CycleGAN. The first kind of development is to enable multi-modal generations: MUNIT~\cite{huang2018munit} and DRIT~\cite{DRIT} decompose the latent space of images into a domain-invariant content space and a domain-specific style space to get diverse outputs. Another enhancement of CycleGAN is to perform translation across multiple (more than two) domains simultaneously, such as StarGAN~\cite{choi2018stargan}. A more funtional line of research focuses on transformation between domains with larger difference. For example,  CoupledGAN~\cite{liu2016coupled}, UNIT~\cite{liu2017unsupervised}, ComboGAN~\cite{anoosheh2018combogan} and XGAN~\cite{royer2017xgan} using domain-sharing latent space, and U-GAT-IT~\cite{Kim2020U-GAT-IT} resort to attention modules for feature selection. Recently, TransGAGA~\cite{wu2019transgaga} and TravelGAN~\cite{amodio2019travelgan} are proposed to characterize the latent representation by using Cartesian product of geometry and preserving vector arithmetic, respectively.


\noindent
\textbf{Introspective Networks.}
Exploring the double roles of the discriminator has been conducted by Introspective Neural Networks (INN)~\cite{jin2017introspective,lazarow2017introspective,lee2018wasserstein} and Introspective Adversarial Networks (IAN)~\cite{brock2017neural,su2019ogan}.
Although INN does share the same purpose of reusing discriminator for generation, it exhibits several remarkable differences compared to our NICE-GAN. First, INN and NICE-GAN tackle different tasks. INN is for pure generation, and the discriminator is reused for generation from hidden vectors to images (as decoding); our NICE-GAN is for translation, and the discriminator is reused for embedding from images to hidden vectors (as encoding). Furthermore, INN requires sequential training even when doing inference, while NICE-GAN only needs one forward pass to generate a novel image, depicting more efficiency. Regarding IAN, it is also for pure generation and reuses one discriminator to generate self-false samples, which is an introspective mechanism; our NICE-GAN reuses the discriminator of one domain to generate a false sample of the other, which is indeed a mutual introspective mechanism.

\begin{figure*}[t]
\setlength{\abovecaptionskip}{-8pt} 
\setlength{\belowcaptionskip}{-10pt} 
\begin{center}
\includegraphics[width=1.0\linewidth]{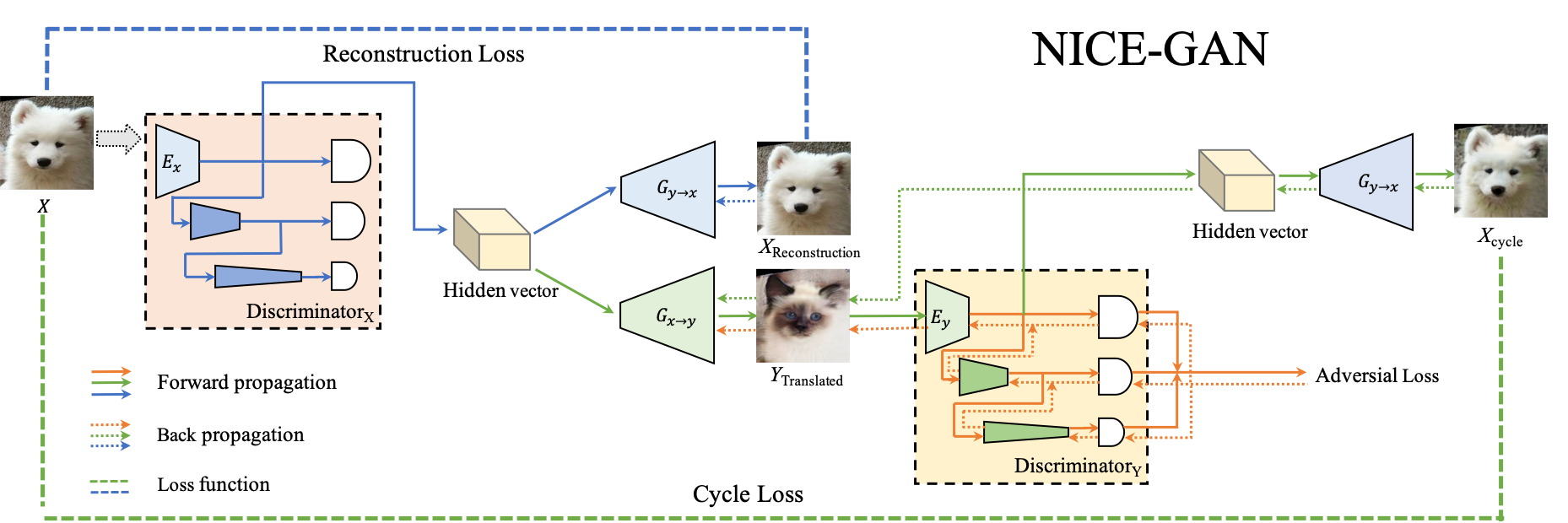}
\end{center}
  \caption{Illustration of the flowchart of NICE-GAN. Here we only display one translation stream from $\mathcal{X}$ to $\mathcal{Y}$ (from dog to cat). Note that we apply a decoupled training fashion: the encoder $E_y$ is fixed when minimizing the adversarial loss, the reconstruction loss and the cycle loss, and it is trained when maximizing the adversarial loss.}
\label{fig:network}
\end{figure*}

\section{Our NICE-GAN}
\vskip -5pt
This section presents the detailed formulation of our method. We first introduce the general idea, and then follow it up by providing the details of each component in NICE-GAN. The decoupled training mechanism is specified as well.

\subsection{General Formulation}
\vskip -3pt
\noindent
\textbf{Problem Definition.}
Let $\mathcal{X}$, $\mathcal{Y}$ be two image domains. While supervised image-to-image translation requires to learn the conditional mappings $f_{x\rightarrow y}=p(\mathcal{Y}|\mathcal{X})$ and $f_{y\rightarrow x}=p(\mathcal{X}|\mathcal{Y})$ given the joint distribution $p(\mathcal{X}, \mathcal{Y})$, unsupervised translation learns $f_{x\rightarrow y}$ and $f_{y\rightarrow x}$ with only the marginals $p(\mathcal{X})$ and $p(\mathcal{Y})$ provided. Unsupervised translation is ill-posed, since there are infinitely many conditional probabilities corresponded to the same marginal distributions. To address this issue, current methods resort to adding extra regulations, such as weight-coupling~\cite{liu2016coupled,liu2017unsupervised,DRIT}, cycle-consistency~\cite{CycleGAN2017,kim2017learning,yi2017dualgan}, and identity-mapping-enforcing~\cite{taigman2016unsupervised,CycleGAN2017}, the latter two of which are employed in this paper. 

In most of existing frameworks, the translation $f_{x\rightarrow y}$ (resp. $f_{y\rightarrow x}$) is composed of an encoder $E_x$ (resp. $E_y$) and a generator $G_{x\rightarrow y}$ (resp. $G_{y\rightarrow x}$). By combining them all together, it gives $y'=f_{x\rightarrow y}(x)=G_{x\rightarrow y}(E_x(x))$ (resp. $x'=f_{y\rightarrow x}(y)=G_{y\rightarrow x}(E_y(y))$).
The GAN~\cite{goodfellow2014generative} training fashion is usually adopted to enable the translated output to fit the distribution of the target domain. Namely, we use a discriminator $D_y$ (resp. $D_x$) to classify between the true image $y$ and the translated image $y'$ (resp. $x$ and $x'$).   

\vskip 3pt
\noindent
\textbf{No Independent Component for Encoding (NICE).}
As mentioned in introduction, our NICE-GAN reuses discriminators for encoding, delivering the advantages of efficiency and effectiveness for training. Formally, we divide the discriminator $D_y$ into the encoding part $E^{D}_y$ and classification part $C_y$. The encoding part $E^{D}_y$ will replace the original encoder in $f_{y\rightarrow x}$, resulting in a new translation $f_{y\rightarrow x}(y)=G_{y\rightarrow x}(E^{D}_y(y))$. Similarly for the discriminator $D_x$, we define $E^{D}_x$ and $C_x$, and reformulate the translation function as $f_{x\rightarrow y}(x)=G_{x\rightarrow y}(E^{D}_x(x))$. As for the classification components $C_x$ and $C_y$, we further employ the multi-scale structure to boost the expressive power. Besides, the newly-formulated encoders $E^{D}_x$ and $E^{D}_y$ exist in the training loops of both translation and discrimination, making them difficult to train. Hence we proposed a decoupled training flowchart in NICE-GAN. The details of the architecture's build-up and training are presented in Section~\ref{sec:architecture} and Section~\ref{sec:training}, respectively. Figure~\ref{fig:network} illustrates our framework. Unless otherwise noticed, \textbf{we will remove the superscript $D$ from $E^{D}_x$ and $E^{D}_y$ for simplicity in what follows}.

\subsection{Architecture}
\label{sec:architecture}
\vskip -3pt
\noindent
\textbf{Multi-Scale Discriminators $D_x$ and $D_y$.}
We only discuss $D_x$ here, since the formulation of $D_y$ is similar. Full details are provided in the supplementary material (SP). Our usage of multi-scale discriminators is inspired from previous works \cite{durugkar2016generative,iizuka2017globally,demir2018patch,wang2018high}. In these approaches, the discriminator of different scale is applied to the image of different size (the small-size images are attained from the original image by down-sampling). In this paper, we consider a more efficient way by regarding the feature maps in different layers of the single input to be the images of different scales, and then feed each of them to the classifier with the corresponding input size for discriminating, which is similar to the application of feature pyramid representations in object detection (\emph{e.g.} SSD~\cite{liu2016ssd} and FPN~\cite{lin2017feature}).

We now introduce our idea in a formal way. As mentioned above, the discriminator $D_x$ contains two parts: the encoder $E_x$ and the classifier $C_x$. To enable multi-scale processing, the classifier $C_x$ is further divided into three sub-classifiers: $C^0_{x}$ for local scale (10 x 10 receptive field), $C^1_{x}$ for middle scale (70 x 70 receptive field), and $C^2_{x}$ for global scale (286 x 286 receptive field). $C^0_{x}$ is directly connected to the output of $E_x$. Then, a down-sampling-convolution layer is conducted on $E_x$ to provide the feature maps of smaller scale, which are concatenated to two branches: one is linked to $C^1_{x}$, and the other one is further down sampled through convolution layers followed by $C^2_{x}$. For a single input image, $C^0_{x}$, $C^1_{x}$, and $C^2_{x}$ are all trained to predict whether the image is true of false. The multi-scale discriminator is also illustrated in Figure~\ref{fig:network}.

Besides the multi-scale design, we develop a residual attention mechanism to further facilitate the feature propagation in our discriminator. Using attention in discriminator is originally proposed by U-GAT-IT~\cite{Kim2020U-GAT-IT}. Suppose the encoder contains feature maps of number $K$ (\emph{i.e.} $E_x=\{E^{k}_x\}_{k=1}^K$). The idea of U-GAT-IT is first learning an attention vector $w$, each element of which counts the importance of each feature map. Then, the attended features computed by $a(x)=w\times E_x(x)=\{w_k\times E^{k}_x(x)\}_{k=1}^K$ are leveraged for later classification. Upon but beyond U-GAT-IT, this paper further takes the residual connection into account, that is, we use $a(x)=\gamma\times w\times E_x(x) + E_x(x)$, where the trainable parameter $\gamma$ determines the trade-off between the attended features and the original ones. When $\gamma=0$, it returns to $E_x(x)$ indicating no attention is used, and otherwise, the attention is activated. By this modification, our method becomes more flexible on adjusting the importance of different feature maps and thus attains more effectiveness in training, which will be clarified by our experiments.

\vskip 3pt
\noindent
\textbf{Generators $G_{x\rightarrow y}$ and $G_{y\rightarrow x}$.}
Both $G_{x\rightarrow y}$ and $G_{y\rightarrow x}$ are composed of six residual blocks~\cite{he2016deep}, and two sub-pixel convolutional layers for up-sampling~\cite{shi2016real}. And, we use the AdaLIN light version similar to the paper~\cite{Kim2020U-GAT-IT}. In addition, spectral normalization~\cite{miyato2018spectral} used for the discriminator and cycle-consistency loss is conducted to prevent generators from mode collapse. Full details are presented in the SP.

\subsection{Decoupled Training}

\label{sec:training}
\vskip -5pt
The training process is proceeded in terms of three kinds of losses: adversarial loss, identity reconstruction loss, and cycle-consistency loss. The adversarial loss is to pursue domain transfer, while both reconstruction loss and cycle-consistency loss are for tackling the non-identifiability issue as pointed out before. 

Since the encoder $E_x$ is not only a part of the discriminator $D_x$ but is also taken as the input of the generator $G_{x\rightarrow y}$, it will incur inconsistency if we apply conventional adversarial training. To overcome this defect, we decouple the training of $E_x$ from that of the generator $G_{x\rightarrow y}$. The details of formulating each loss are provided below.

\noindent
\textbf{Adversarial loss.}
First, we make use of the least-square adversarial loss by~\cite{mao2017least} for its more stable training and higher quality of generation. The min-max game is conducted by

\begin{eqnarray}
\nonumber
 \min_{G_{x\rightarrow y}}\max_{D_y=(C_y\circ E_y)} L^{x\rightarrow y}_{gan} := \mathbb{E}_{y\sim \mathcal{Y}}\left[\left(D_y(y)\right)^2\right]\notag\\
+ \mathbb{E}_{x\sim \mathcal{X}}\left[\left(1-D_y(G_{x\rightarrow y}(E_x(x)))\right)^2\right], 
\end{eqnarray}
where, $E_x$ is fixed and $E_y$ is trained when maximizing $L^{x\rightarrow y}_{gan}$, and both of them are fixed when minimizing $L^{x\rightarrow y}_{gan}$.

\vskip 3pt
\noindent
\textbf{Cycle-consistency loss.}
The cycle-consistency loss is first introduced by CycleGAN~\cite{CycleGAN2017} and DiscoGAN~\cite{kim2017learning}, which is to force the generators to be each other’s inverse. 
\begin{align}
\min_{\substack{G_{x\rightarrow y}\\ G_{y\rightarrow x}}}  L^{x\rightarrow y}_{cycle} := \mathbb{E}_{x\sim \mathcal{X}}\left[\left|x - G_{y\rightarrow x}(E_y(G_{x\rightarrow y}(E_x(x))))\right|_1\right],
\end{align}
where $|\cdot|_1$ computes the $\ell_1$ norm, and both $E_x$ and $E_y$ are also frozen.

\noindent
\textbf{Reconstruction loss.}
Forcing the generator to be close to the identity function is another crucial regulation technique in CycleGAN~\cite{CycleGAN2017}. Unlike CycleGAN where the identity loss is based on domain similarity assumption, our reconstruction is based on the shared-latent space assumption. Reconstruction loss is to regularize the translation to be near an identity mapping when real samples' hidden vectors of the source domain are provided as the input to the generator of the source domain. Namely,
\begin{eqnarray}
\min_{G_{y\rightarrow x}} L^{x\rightarrow y}_{recon} := \mathbb{E}_{x\sim \mathcal{X}}\left[\left|x - G_{y\rightarrow x}(E_x (x))\right|_1\right],
\end{eqnarray}
where $E_x$ is still kept unchanged.

Similarly, we can define the losses from domain $\mathcal{Y}$ to $\mathcal{X}$: $L^{y\rightarrow x}_{gan}$, $L^{y\rightarrow x}_{cycle}$, and $L^{y\rightarrow x}_{recon}$. 

\begin{figure*}[t]
\setlength{\abovecaptionskip}{-15pt} 
\setlength{\belowcaptionskip}{-10pt} 
\begin{center}
\includegraphics[width=0.82\linewidth]{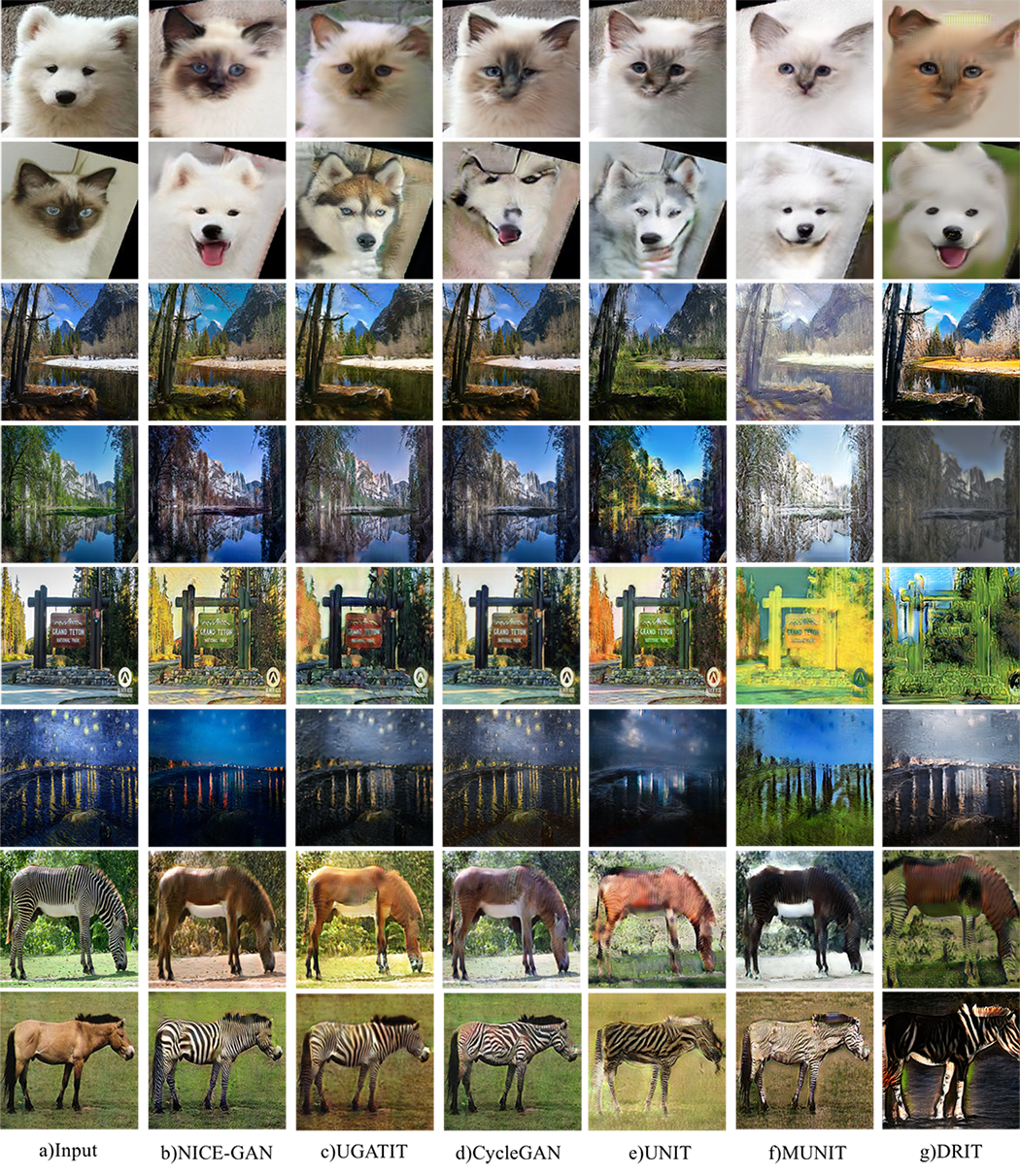}
\end{center}
   \caption{{\bf Examples of generated outputs.} From top to bottom:  dog$\leftrightarrow$cat, winter$\leftrightarrow$summer, photo$\leftrightarrow$vangogh, and zebra$\leftrightarrow$horse. }
\label{fig:contrast}
\end{figure*}

\vskip 3pt
\noindent
\textbf{Full objective.}
The discriminators' final objective is
\begin{align}
\max_{E_x, C_x, E_y, C_y} \lambda_1 L_{gan}; 
\end{align}

while the generators' final loss objective is
\begin{align}
&\min_{G_{x\rightarrow y},G_{y\rightarrow x}}  \notag
\lambda_1 L_{gan} + \lambda_2 L_{cycle} + \lambda_3 L_{recon},
\end{align} 
where, $L_{gan} = L^{x\rightarrow y}_{gan} +  L^{y\rightarrow x}_{gan}, L_{cycle} = L^{x\rightarrow y}_{cycle} +  L^{y\rightarrow x}_{cycle}, L_{recon} = L^{x\rightarrow y}_{recon} +  L^{y\rightarrow x}_{recon}$, and $\lambda_1$, $\lambda_2$, and $\lambda_3$ are the trade-off weights (they are fixed as $\lambda_1=1$, $\lambda_2=10$, and $\lambda_3=10$ throughout our experiments). 

Note again, the encoders $E_x$ and $E_y$ are trained under the discriminator's objective but are decoupled from the training of generators. The benefit of the proposed decoupled training paradigm will be analyzed by our experiments.

\begin{table*}[t]
\setlength{\abovecaptionskip}{-10pt} 
\setlength{\belowcaptionskip}{-10pt} 
\caption{The FID and the KID $\times$100 for different algorithms. Lower is better. All of the methods are trained to the 100K-th iterations. NICE-GAN* is the version that the generator network is composed of only four residual blocks.}
\begin{center}

  \label{tab:FID_KID}
    \begin{tabular}{|l|c|c|c|c|c|c|c|c|}
    \hline
    \multirow{2}{*}{\diagbox{Method}{Dataset}}& 
    \multicolumn{2}{c|}{dog $\rightarrow$ cat}&\multicolumn{2}{c|}{winter $\rightarrow$ summer} &\multicolumn{2}{c|}{photo $\rightarrow$ vangogh} &\multicolumn{2}{c|}{zebra $\rightarrow$ horse}\cr\cline{2-9}
      & FID & KID $\times$ 100 & FID & KID $\times$ 100 & FID & KID $\times$ 100 & FID & KID $\times$ 100 \cr
    \hline \hline
      NICE-GAN   & {\bf48.79}	&1.58	&{\bf76.44}	&{\bf1.22}	&{\bf122.27}	& 3.71	&149.48	 &4.29 \cr\cline{1-9}
      NICE-GAN*   &51.98	&1.50	&79.02	&1.35	&122.59	&{\bf3.53}	&150.57	&4.43 \cr\cline{1-9}
       U-GAT-IT-light &80.75 &3.22 &80.33 &1.82 &137.70 &6.03 &{\bf145.47} &{\bf3.39} \cr\cline{1-9}
       CycleGAN & 119.32 & 4.93 & 79.58 &1.36 &	136.97 &	4.75 &156.19 &5.54 \cr\cline{1-9}
       UNIT  &59.56	& 1.94 &	95.93 &	4.63 &136.80 &5.17 &170.76 &6.30 \cr\cline{1-9}
        MUNIT   &53.25	&{\bf1.26}	&99.14	&4.66	&130.55	&4.50	&193.43	&7.25\cr\cline{1-9}
        DRIT   &94.50	&5.20	&78.61	&1.69	&136.24	&5.43	&200.41	&10.12\cr\cline{1-9}
    \hline \hline 
    \multirow{2}{*}{\diagbox{Method}{Dataset}}& 
    \multicolumn{2}{c|}{cat $\rightarrow$ dog}&\multicolumn{2}{c|}{summer $\rightarrow$ winter } &\multicolumn{2}{c|}{vangogh $\rightarrow$ photo} &\multicolumn{2}{c|}{horse $\rightarrow$ zebra}\cr\cline{2-9}
      & FID & KID $\times$ 100 & FID & KID $\times$ 100 & FID & KID $\times$ 100 & FID & KID $\times$ 100 \cr
    \hline \hline
    NICE-GAN	&{\bf44.67}	&{\bf1.20}	&{\bf76.03}	&{\bf0.67}	&{\bf112.00}	&{\bf2.79}	&{\bf65.93}	&{\bf2.09}  \cr\cline{1-9}
    NICE-GAN*	&55.72	&1.89	&77.13	&0.73	&117.81	&3.61	&84.89	&3.29  \cr\cline{1-9}
    U-GAT-IT-light	&64.36	&2.49	&88.41	&1.43	&123.57	&4.91	&113.44	&5.13 \cr\cline{1-9}
    CycleGAN	&125.30	&6.93	&78.76	&0.78	&135.01	&4.71	&95.98	&3.24 \cr\cline{1-9}
    UNIT	&63.78	&1.94	&112.07	&5.36	&143.96	&7.44	&131.04	&7.19 \cr\cline{1-9}
    MUNIT	&60.84	&2.42	&114.08	&5.27	&138.86	&6.19	&128.70	&6.92 \cr\cline{1-9}
    DRIT	&79.57	&4.57	&81.64	&1.27	&142.69	&5.62	&111.63	&7.40 \cr\cline{1-9}
    \end{tabular}
\end{center}
\end{table*}

\section{Experiments}
\subsection{Baselines}
\vskip -5pt
We compare the performance NICE-GAN with state-of-the-art methods including CycleGAN~\cite{CycleGAN2017}, UNIT~\cite{liu2017unsupervised}, MUNIT~\cite{huang2018munit}, DRIT~\cite{DRIT}, and U-GAT-IT~\cite{Kim2020U-GAT-IT} considering their competitive performance on unsupervised image-to-image translation. All compared methods are conducted by using the public codes. Specifically for U-GAT-IT, we use its light version due to the memory limit of our GPU machine. The details of all baselines are introduced in the SP.

\subsection{Dataset}
\vskip -5pt
The experiments are carried out on four popular benchmarks of unpaired images: {\bf horse$\leftrightarrow$zebra}, {\bf summer$\leftrightarrow$winter\_yosemite}, {\bf vangogh$\leftrightarrow$photo} and {\bf cat$\leftrightarrow$dog}. The first three datasets are used in CycleGAN~\cite{CycleGAN2017}, whose train-test splits are respectively: 1,067/120 (horse), 1,334/140 (zebra); 1,231/309 (summer), 962/238 (winter); 400/400 (vangogh), 6,287/751 (photo). The last dataset is studied in DRIT~\cite{DRIT}\cite{DRIT_plus}, whose train-test splits are: 771/100 (cat), 1,264/100 (dog). All images of all datasets are cropped and resized to 256 $\times$ 256 for training and testing.

\begin{table}[t]
\vskip -8pt
\setlength{\abovecaptionskip}{-10pt} 
\setlength{\belowcaptionskip}{-10pt} 
\caption{Total number of parameters and FLOPs of network modules. NICE-GAN* are the version that the generator network is composed of only four residual blocks. }
\begin{center}

    \label{tab:Total number of parameters}
    \begin{tabular}{|l|c|c|}
    \hline
    \multirow{2}{*}{\diagbox{Method}{Module}}& 
    \multicolumn{2}{c|}{Total number of params(FLOPs)}\cr\cline{2-3}
      & Generators & Discriminators  \cr
    \hline \hline
      U-GAT-IT-light &21.2M(105.0G)	&112.8M(15.8G) \cr\cline{1-3}
      NICE-GAN &16.2M(67.6G)	&93.7M(12.0G) \cr\cline{1-3}
      NICE-GAN* &11.5M(48.2G)	&93.7M(12.0G)  \cr\cline{1-3}
    \end{tabular}
\end{center}
\end{table}

\subsection{Evaluation Metrics}
\vskip -3pt
\noindent
{\bf Human Preference.} To compare the veracity of translation outputs generated by different methods, we carry out human perceptual study. Similar to Wang \etal~\cite{wang2018high}, volunteers are shown an input image and three translation outputs from different methods, and given unlimited time they select which translation output looks better. 

\noindent
{\bf The Fréchet Inception Distance (FID)}  proposed by Heusel \etal (2017) \cite{heusel2017gans} contrast the statistics of generated samples against real samples. The FID fits a Gaussian distribution to the hidden activations of InceptionNet for each compared image set and then computes the Fréchet distance (also known as the Wasserstein-2 distance) between those Gaussians. Lower FID is better, corresponding to generated images more similar to the real. 

\begin{figure}[t]
\vskip -15pt
\setlength{\abovecaptionskip}{-7pt} 
\setlength{\belowcaptionskip}{10pt} 
\begin{center}
\includegraphics[width=1.0\linewidth]{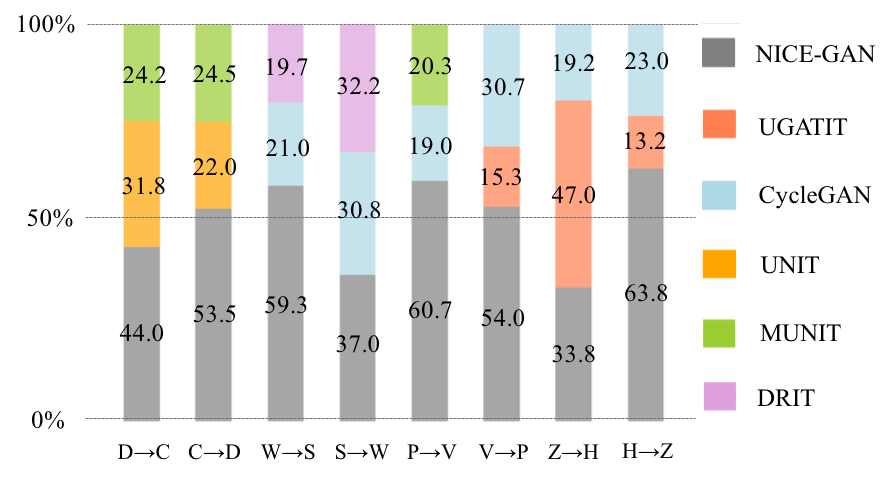}
\end{center}
   \caption{ {\bf Human preference results.} The number indicates the percentage of preference on that translation task. Abbreviation: (D)og, (C)at; (W)inter, (S)ummer; (P)hoto, (V)angogh; (Z)ebra, (H)orse.}
\label{fig:column_big}
\end{figure}

\noindent
{\bf The Kernel Inception Distance (KID)} developed by~\cite{binkowski2018demystifying} is a metric similar to the FID but uses the squared Maximum Mean Discrepancy(MMD) between Inception representations with a polynomial kernel, $k(x, y) = \left(\frac{1}{d}x^{\rm T}y + 1\right)^3$, where $d$ is the representation dimension. It can also be viewed as an MMD directly on input images with the kernel $K(x, y) = k(\theta(x), \theta(y))$, where $\theta$ is the function mapping images to Inception representations.  Unlike FID, KID has a simple unbiased estimator, making it more reliable especially when there are much more inception features channels than image numbers. Lower KID indicates more visual similarity between real and generated images. Our implementation of KID is based on~\href{https://github.com/mbinkowski/MMD-GAN}{https://github.com/mbinkowski/MMD-GAN} where the hidden representations are from the Inception-v3 pool3 layer.



\subsection{Setup}
\vskip -5pt
 
We use ReLU as the actionvation function in the generator and leaky-ReLU with a slope of $0.2$ in the discriminator. We train all models using the Adam~\cite{kingma2014adam} optimizer with the learning rate $0.0001$ and $(\beta 1 , \beta 2 ) = (0.5, 0.999)$ on NVIDIA RTX 2080Ti GPUs. For data augmentation, we flip the images horizontally with a probability of $0.5$, resized them to $286 \times 286$, and randomly cropped them to $256 \times 256$. The batch size is set to $1$ for all experiments. We also use a weight decay at the rate of $0.0001$. All models are trained over 100K iterations. More details on the training process and network architecture are provided in the SP.

\subsection{Comparisons with state of the arts}
\vskip -5pt
Table~\ref{tab:FID_KID} shows that our approach generally achieves the lowest FID or KID scores on all cases except zebra$\rightarrow$horse, indicating the promising translation ability of our NICE framework on varying tasks. These two metrics maintain good consistency in relative scores, which fully demonstrates our NICE-GAN reasonably performs well regardless of what measure we have used. By contrast, other methods only perform well on certain datasets; for instance, U-GAT-IT-light, UNIT and MUNIT successfully transforms the semantic of objects(\emph{e.g.} animal faces), while CycleGAN is good at modifying low-level characteristics (\emph{e.g.} colors and textures).
U-GAT-IT-light roughly shares the same structures (multi-scale discriminators and generators) as NICE-GAN, and it differs from NICE-GAN mainly in its independent formulation of encoders. Table~\ref{tab:Total number of parameters} reports total number of parameters and FLOPs of U-GAT-IT-light and NICE-GAN, and it reads that 
our architecture is more compact by reusing discriminators for encoding. To further observe the visual difference, Figure~\ref{fig:contrast} depicts the translated images of different methods on test sets. The generated images by NICE-GAN are almost more meaningful and have less artifacts than others (see the cat$\leftrightarrow$dog task for an example).

In addition to our method, we select two baselines achieving lowest KID scores in each dataset to conduct a human perceptual study. Firstly, volunteers are shown an example pair consisting of source-domain image and a target-domain image as a reference to better understand what style is translating. Secondly, they are given an input image, and three translated outputs among which one is from NICE-GAN and the other two from the selected baselines. They have unlimited time to choose which looks most real based on perceptual realism. The synthesized images are displayed in a randomized order to ensure fair comparisons. Besides, checkpoint questions are set and distributed to each volunteer to validating human effort.  A total of 123 questionnaires are collected in which we find that 120 are valid. Figure~\ref{fig:column_big} shows that NICE-GAN wins the majority of votes in all cases except for zebra$\rightarrow$horse. These results are also consistent with the quantitative metrics in Table~\ref{tab:FID_KID}. More examples of the results from our model are included in the SP. 
\begin{table}[t]
\vskip -5pt
\setlength{\abovecaptionskip}{-7pt} 
\setlength{\belowcaptionskip}{-7pt} 
\caption{{\bf Ablation Study.} Results of methods are all in 100K iterations of discriminator.  NICE: No Independent Component for Encoding; RA: add residual connection in CAM attention module; $C^0_{x}$ for local scale (10 x 10 receptive field), $C^1_{x}$ for middle scale (70 x 70 receptive field), and $C^2_{x}$ for global scale (286 x 286 receptive field); $-$: decreasing the number of shared layers by 1; $+$: increasing  by 1.}
\begin{center}
\begin{tabular}{|l|lllll|l|l|}
\hline
Data & \multicolumn{5}{c|}{Components} & \multirow{2}{*}{FID} & {KID} \\  \cline{2-6}
Set                        & NICE    & RA   & $C^0_{x}$ & $C^1_{x}$ & $C^2_{x}$   &             &         $\times$ 100                 \\ \hline \hline
                       & $\times$       & $\times$   & \checkmark   & \checkmark   & \checkmark   & 80.75      & 3.22      \\ 
                                             & $\times$       & \checkmark   & \checkmark   & \checkmark   & \checkmark   & 67.60                   & 2.94     \\
                                    & \checkmark       & $\times$   & \checkmark   & \checkmark   & \checkmark   & 63.80                    & 3.27                        \\ 
       dog     & $-$       & \checkmark   & \checkmark   & \checkmark   & \checkmark   & {\bf 48.55}                    & {\bf 1.23}                        \\
   $\rightarrow$      & \checkmark       & \checkmark   & \checkmark   & \checkmark   & \checkmark   &  48.79                   &  1.58                        \\ 
     cat     & $+$        & \checkmark   & \checkmark   & \checkmark   & \checkmark   & 53.52                     &  1.84                        \\ 
               & \checkmark       & \checkmark   & \checkmark    & \checkmark   & $\times$   & 203.56                    & 15.27                        \\
                                               & \checkmark       & \checkmark   & \checkmark    & $\times$    & $\times$   & 216.03                    & 18.57                        \\ \hline \hline
                        & $\times$       & $\times$   & \checkmark   & \checkmark   & \checkmark   & 64.36                    & 2.49                        \\ 
                                               & $\times$       & \checkmark   & \checkmark   & \checkmark   & \checkmark   & 64.62                    & 2.41                        \\ 
                                              & \checkmark       &  $\times$   & \checkmark   & \checkmark   & \checkmark   & 51.49                   & 1.68                        \\  
                                   cat & $-$       & \checkmark   & \checkmark   & \checkmark   & \checkmark   &  52.92                   &  1.82                      \\ 
                                    $\rightarrow$           & \checkmark       & \checkmark   & \checkmark   & \checkmark   & \checkmark   & {\bf 44.67}                    & {\bf 1.20}                       \\
                                    dog & $+$       & \checkmark   & \checkmark   & \checkmark   & \checkmark   &  54.90                   &  2.17                       \\ 
                                               & \checkmark       & \checkmark   & \checkmark    & \checkmark   & $\times$   & 238.62                    & 21.41                        \\
                                               & \checkmark       & \checkmark   & \checkmark    & $\times$    & $\times$   & 231.24                   & 22.12                        \\ \hline 
\end{tabular}
\label{tab:ablation1}
\end{center}
\end{table}

\subsection{Ablation study}
\vskip -5pt
We conduct ablation studies on the cat$\leftrightarrow$dog datasets in Table~\ref{tab:ablation1} to isolate the validity of the key components of our method: the NICE strategy, the multi-scale formulation and the Residual Attention (RA) mechanism in discriminators. We perform four groups of experiments. 

\begin{figure}[t]
\vskip -12pt
\setlength{\abovecaptionskip}{-7pt} 
\setlength{\belowcaptionskip}{5pt} 
\begin{center}
\includegraphics[width=1.0\linewidth]{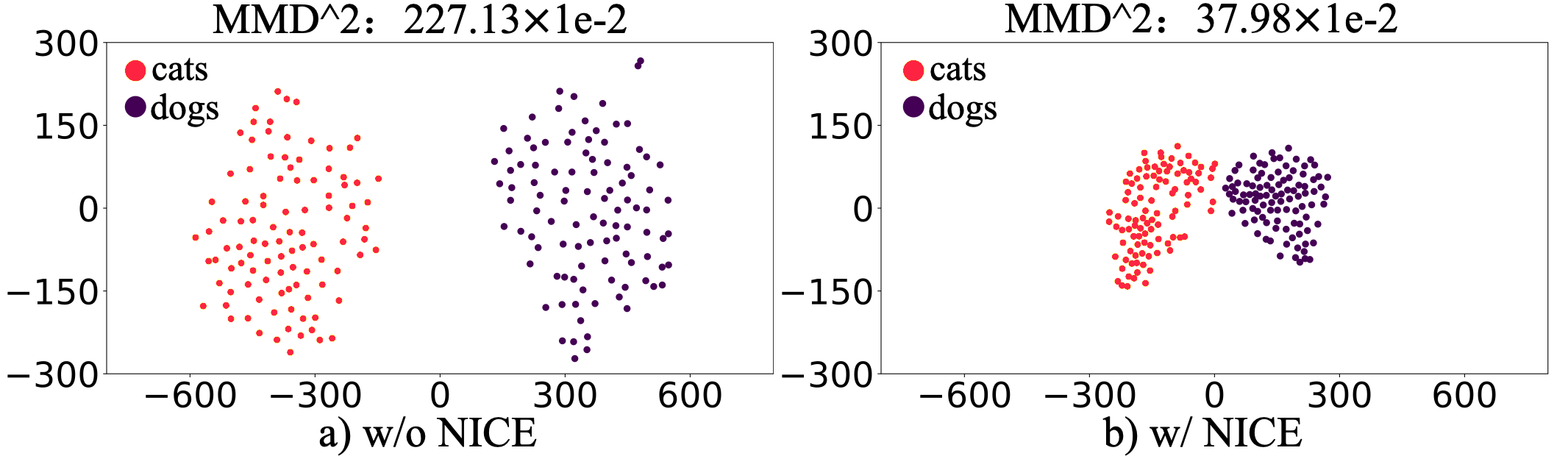}
\end{center}
\caption{The t-SNE visualization of the latent vectors, as well as the MMD to measure domain difference. }
\label{fig:tsne}
\end{figure}

\noindent
\textbf{NICE and RA.}
The first group keeps employing the multi-scale formulation but removes either or both of NICE and RA to draw the performance difference. The results are reported in Table~\ref{tab:ablation1}. It verifies that each of NICE and RA contributes to the performance improvement while the importance of NICE is more significant. Overall, by combing all components, NICE-GAN remarkably outperforms all other variants. \autoref{fig:tsne} shows the latent vectors of each domain w/ and w/o NICE on \emph{cat $\leftrightarrow$dog }  via t-SNE, as well as MMD to compute domain difference. Interestingly, with NICE training, the latent distributions of two domains become more clustered and closer, yet separable to each other. Such phenomenon explains why our NICE-GAN performs promisingly. By shortening the transition path between domains in the latent space, NICE-GAN built upon the shared latent space assumption can probably facilitate domain translation in the image space.

\noindent
\textbf{The number of shared layers.}
For consistent comparison, we employ the commonly-used ResNet backbone in~\cite{CycleGAN2017} as our generator and naturally share the whole encoder therein. We also evaluate the effect of changing the number of layers shared by discriminator and encoder. 
Table~\ref{tab:ablation1} shows that whether decreasing or increasing the number of layers generally hinders the performance.
Thus, the best choice here is sharing the whole default encoder. 


\noindent
\textbf{Multi-scale.} 
The third group of experiments is to evaluate the impact of the multi-scale formulation ($C^0_x$, $C^1_x$, $C^2_x$) in discriminators. The results are summarized in Table~\ref{tab:ablation1}. We find that removing $C^2_x$ will cause a serious detriment; the importance of $C^0_x$ and $C^1_x$ is task-dependent, and adding $C^1_x$ upon  $C^0_x$ does not exhibit clear enhancement on this task. Actually, all three scales are generally necessary and multi-scale is more robust, with more discussions in the SP.

\noindent
\textbf{Weight-coupling.}
Besides, there are existing methods of model compression, such as weight-coupling~\cite{liu2017unsupervised,DRIT}.  Sharing a few layers between two generators enables model compression but detriments the translation performance. For example on \emph{cat$\leftrightarrow$dog}, the FID increases from 48.79/44.67 to 49.67/56.32 if sharing the first layer of the decoders, and from 48.79/44.67 to 55.00/55.60 if sharing the last layer of encoders. Similar results are observed if we reuse the first layer of the classifiers, the FID increases from 48.79/44.67 to 61.73/46.65 .
It implies weight-coupling could weaken the translation power for each domain.

\subsection{Decoupled Training Analysis}
\label{sec:training_exp}
\vskip -5pt

In our NICE framework, we decouple the training of $E_x$ from that of the generator $G_{x\rightarrow y}$. To prove the effectiveness of this strategy, we develop two additional variants: NICE-GAN-1 and NICE-GAN-2. To be specific, NICE-GAN-1 adopts the conventional training approach, where the encoders are jointly trained with the discriminators and generators. As for NICE-GAN-2, it is also performed in a decoupled way but acts oppositely with our method, that is, the encoders are trained along with the generator, independent to the classifiers in the discriminators. From a more essential sense, the discriminators indeed degenerate to the classifiers in NICE-GAN-2.



\begin{figure}[t]
\vskip -16pt
\setlength{\abovecaptionskip}{-7pt} 
\setlength{\belowcaptionskip}{5pt} 
\begin{center}
\includegraphics[width=1.0\linewidth]{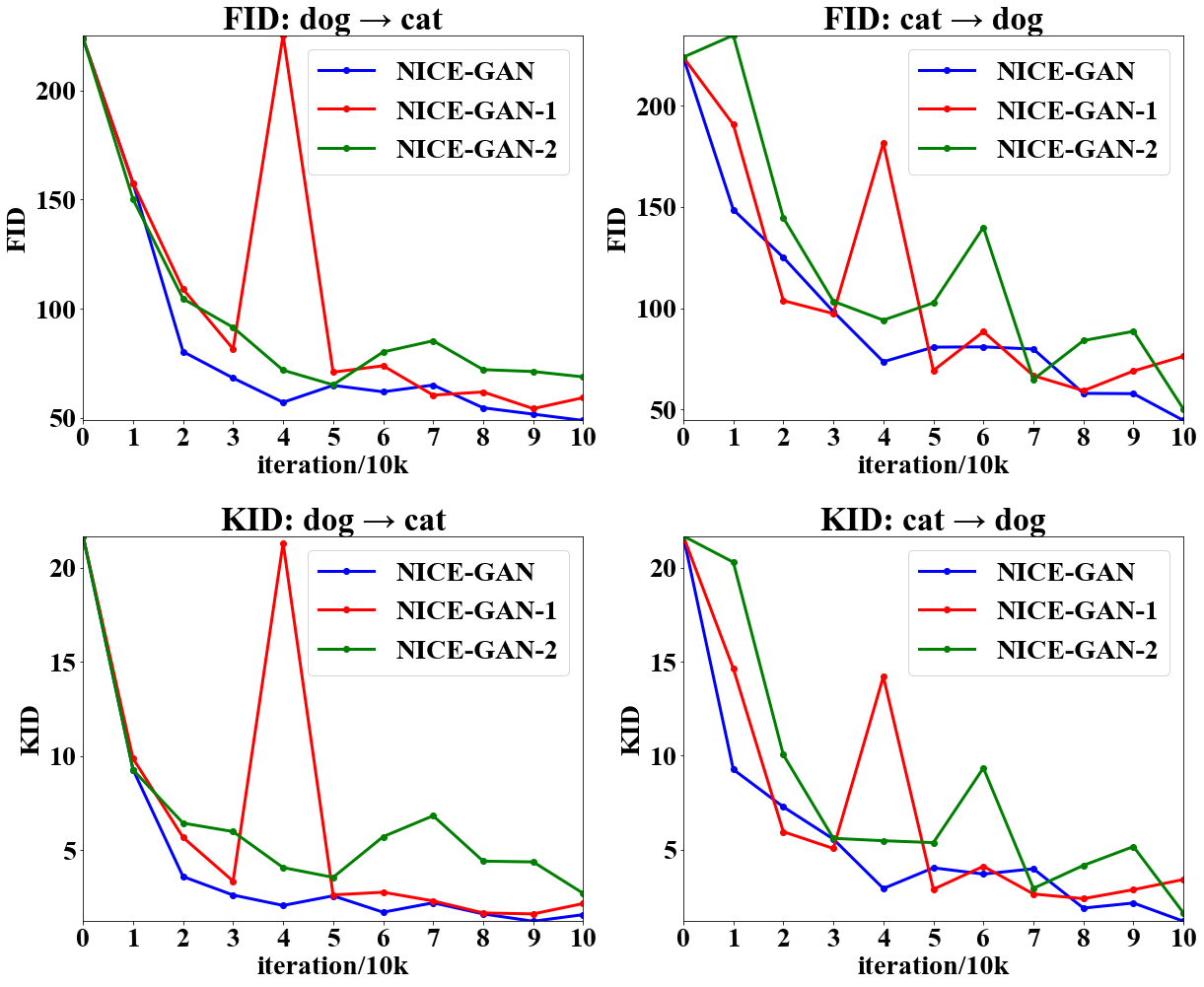}
\end{center}
\caption{{\bf Decoupled Training Analysis.} NICE-GAN: decoupled training, $E_x$ and $E_y$ will only be updated by $max L_{gan}$; NICE-GAN-1: jointly train the discriminators and generators, $E_x$ and $E_y$ will be updated by $min max L_{gan}$,$min L_{cycle}$and $min L_{recon}$ ; NICE-GAN-2: decoupled training, $E_x$ and $E_y$ will be updated by $min L_{gan}$,$min L_{cycle}$and $min L_{recon}$. }
\label{fig:zxt}
\end{figure}

Figure~\ref{fig:zxt} reports the training curves of NICE-GAN, NICE-GAN-1 and NICE-GAN-2. Clearly, the training of NICE-GAN-1 is unstable, which is consistent with our analysis. NICE-GAN-2 performs more stably and better than NICE-GAN-1, but is still inferior to our NICE-GAN. We conjecture that in NICE-GAN-2, using the classifiers for discriminating is actually aligning the distributions of hidden vectors. Nevertheless, NICE-GAN leverages both the encoders and  classifiers for discriminating, underlying that it is matching the distributions of image space, thus more precise information is captured.

A clear disentanglement of responsibilities of different components makes NICE-GAN simple and effective. Besides, It further supports the idea~\cite{brock2017neural} that features learned by a discriminatively trained network tend to be more expressive than those learned by an encoder network trained via maximum likelihood, and thus better suited for inference.

\section{Conclusion}
\vskip -5pt
In this paper, we present NICE-GAN, a novel framework for unsupervised image-to-image translation. It reuses discriminators for encoding and develops a decoupled paradigm for efficient training. Comparable experimental evaluations on several popular benchmarks reveal that NICE-GAN generally achieves superior performance over state-of-the-art methods. Our research is expected to evoke the rethinking on what discriminators actually can do, and it is potentially applicable to refresh the GAN-based models in other cases.

\vskip 3pt
\noindent
\textbf{Acknowledgement.}
This research was jointly funded by the National Natural Science Foundation of China (NSFC) and the German Research Foundation (DFG) in project Cross Modal Learning, NSFC 61621136008/DFG TRR-169, and the National Natural Science Foundation of China(Grant No.91848206). We thank Chengliang Zhong, Mingxuan Jing and Dr. Tao Kong for the insightful advice.

{\small
\bibliographystyle{ieee_fullname}
\bibliography{egbib}
}

\clearpage
\begin{appendix}
\section{Appdendix}
\subsection{Introduction of state-of-the-art models}

{\bf CycleGAN} uses an adversarial loss to learn the mapping between two different domains. The method regularizes the mapping through cycle-consistency losses, using two down-sampling convolution blocks, nine residual blocks, two up-sampling deconvolution blocks and four discriminator layers.
Codes are on \href{https://github.com/junyanz/pytorch-CycleGAN-and-pix2pix}{https://github.com/junyanz/pytorch-CycleGAN-and-pix2pix}.

{\bf UNIT} consists of two VAE-GAN with shared latent space. The structure of the UNIT is similar to CycleGAN, but different from CycleGAN in that it uses multi-scale discriminators and shares the weight of the high-level layer stage of the encoder and decoder.
Codes are on \href{https://github.com/mingyuliutw/UNIT}{https://github.com/mingyuliutw/UNIT}.

{\bf MUNIT} can generate various outputs for a single input image. MUNIT assumes that the image representation can be decomposed into a content code and a style code. The main difference between MUNIT’s network structure and other networks is that it uses AdaIN in the decoder and also a multi-scale discriminator. We generate N = 1 images for each input image in the test set. We use the generated samples and all samples in test set to compute FID and KID.
Codes are on \href{https://github.com/NVlabs/MUNIT}{https://github.com/NVlabs/MUNIT}.

{\bf DRIT}  can also create different outputs for a single input image similar to MUNIT. It decomposes the image into a content code and a style code, using a multi-scale discriminator. The difference between DRIT and MUNIT is that the content code is shared like UNIT. We generate N = 1 images for each input image in the test set. We use the generated samples and all samples in test set to compute FID and KID.
Codes are on \href{https://github.com/HsinYingLee/DRIT}{https://github.com/HsinYingLee/DRIT}.

{\bf U-GAT-IT} is a recent work associated with unsupervised image-to-image translation, which incorporates a CAM (Class Activation Map) module and an AdaLIN (Adaptive Layer-Instance Normalization) function in an end-to-end manner. U-GAT-IT can translate images requiring holistic changes or  large shape changes. Light version is applied due to the limited memory of our gpu. Codes are on \href{https://github.com/znxlwm/UGATIT-pytorch}{https://github.com/znxlwm/UGATIT-pytorch}.

\subsection{Network Architecture}

The architectures of the discriminator and generator in NICE-GAN are shown in Table~\ref{tab:Discriminator} and ~\ref{tab:Generator}, respectively. For the generator network, we use adaptive layer-instance normalization in decoders except the last output layer. For the discriminator network, Leaky-ReLU is applied with a negative slope of 0.2 and spectral normalization is put in all layers. We apply \emph{softmax} instead of \emph{clip} to limit $\rho \in [0,1]$ in AdaLIN. Besides, we concat global average \& max pooling's feature maps before Classifier0 so that the input channel of MLP-(N1) is 256. More details are presented in our source code.
There are some notations: N is the number of output channels; K is the kernel size; S is the side size; P is the padding size; AdaLIN is the adaptive layer-instance normalization; LIN is layer-instance normalization; SN is the spectral normalization; RA is adding residual connection in CAM attention module.

\begin{table*}[!htp]
\caption{Discriminator network architecture}
\begin{center}
\label{tab:Discriminator}
\begin{tabular}{|l|c|c|}
\hline
    Component & Input $\rightarrow $ Output Shape & Layer Information  \\ \hline \hline
Encoder  & $(h, w, 3) \rightarrow (\frac{h}{2} , \frac{w}{2} , 64)$  & CONV-(N64, K4, S2, P1), SN, Leaky-ReLU \\ \cline{2-3} 
  Down-sampling0  & $(\frac{h}{2} , \frac{w}{2} , 64) \rightarrow (\frac{h}{4} , \frac{w}{4} , 128)$  & CONV-(N128, K4, S2, P1), SN, Leaky-ReLU  \\ \hline
RA of  & \multirow{2}{*}{$(\frac{h}{4} , \frac{w}{4} , 128) \rightarrow (\frac{h}{4} , \frac{w}{4} , 256)$} & Global Average \& Max Pooling,  \\  
     Encoder\&            &                   & MLP-(N1), Multiply the weights of MLP \\ \cline{2-3} 
     Classifier0             &    $(\frac{h}{4} , \frac{w}{4} , 256) \rightarrow (\frac{h}{4} , \frac{w}{4} , 128)$              & CONV-(N128, K1, S1), RA, Leaky-ReLU     \\ \hline
                  
  Down-sampling1  & $(\frac{h}{4} , \frac{w}{4} , 128) \rightarrow (\frac{h}{8} , \frac{w}{8} , 256)$   & CONV-(N256, K4, S2, P1), SN, Leaky-ReLU   \\ \hline                  
\multirow{2}{*}{Classifier1} &  $(\frac{h}{8} , \frac{w}{8} , 256) \rightarrow (\frac{h}{8}-1 , \frac{w}{8}-1 , 512)$   & CONV-(N512, K4, S1, P1), SN, Leaky-ReLU   \\ \cline{2-3} 
                  & $(\frac{h}{8}-1 , \frac{w}{8}-1 , 512) \rightarrow (\frac{h}{8}-2 , \frac{w}{8}-2 , 1)$ &  CONV-(N1, K4, S1, P1), SN \\ \hline
\multirow{2}{*}{Down-sampling2} & $(\frac{h}{8} , \frac{w}{8} , 256) \rightarrow (\frac{h}{16} , \frac{w}{16} , 512)$ &  CONV-(N512, K4, S2, P1), SN, Leaky-ReLU \\ \cline{2-3} 
                  & $(\frac{h}{16} , \frac{w}{16} , 512) \rightarrow (\frac{h}{32} , \frac{w}{32} , 1024)$ & CONV-(N1024, K4, S2, P1), SN, Leaky-ReLU   \\ \hline
\multirow{2}{*}{Classifier2} & $(\frac{h}{32} , \frac{w}{32} , 1024) \rightarrow (\frac{h}{32}-1 , \frac{w}{32}-1 , 2048)$  & CONV-(N2048, K4, S1, P1), SN, Leaky-ReLU  \\ \cline{2-3} 
                  & $(\frac{h}{32}-1 , \frac{w}{32}-1 , 2048) \rightarrow (\frac{h}{32}-2 , \frac{w}{32}-2 , 1)$  & CONV-(N1, K4, S1, P1), SN \\ \hline
\end{tabular}
\end{center}
\end{table*}

\begin{table*}[!htpb]
\caption{Generator network architecture}
\begin{center}
\label{tab:Generator}
\begin{tabular}{|l|c|c|}
\hline
    Component & Input $\rightarrow $ Output Shape & Layer Information  \\ \hline \hline
  Sampling  & $ (\frac{h}{4} , \frac{w}{4} , 128)\rightarrow (\frac{h}{4} , \frac{w}{4} , 256)$  &  CONV-(N256, K3, S1, P1), LIN, ReLU \\ \hline
\multirow{4}{*}{$\gamma_{AdaLIN}, \beta_{AdaLIN}$} & $ (\frac{h}{4} , \frac{w}{4} , 256)\rightarrow (1 , 1 , 256)$ & Global Average  Pooling \\ \cline{2-3} 
                  & $ (1 , 1 , 256)\rightarrow (1 , 1 , 256)$ & MLP-(N256), ReLU  \\ \cline{2-3} 
                  & $ (1 , 1 , 256)\rightarrow (1 , 1 , 256)$ & MLP-(N256), ReLU \\ \cline{2-3} 
                  & $ (1 , 1 , 256)\rightarrow (1 , 1 , 256)$ & MLP-(N256), ReLU \\ \hline
\multirow{6}{*}{Bottleneck } &$ (\frac{h}{4} , \frac{w}{4} , 256)\rightarrow (\frac{h}{4} , \frac{w}{4} , 256)$  & AdaResBlock-(N256, K3, S1, P1), AdaLIN, ReLU  \\ \cline{2-3} 
                  &$ (\frac{h}{4} , \frac{w}{4} , 256)\rightarrow (\frac{h}{4} , \frac{w}{4} , 256)$   &AdaResBlock-(N256, K3, S1, P1), AdaLIN, ReLU   \\ \cline{2-3} 
                  &$ (\frac{h}{4} , \frac{w}{4} , 256)\rightarrow (\frac{h}{4} , \frac{w}{4} , 256)$   &AdaResBlock-(N256, K3, S1, P1), AdaLIN, ReLU   \\ \cline{2-3} 
                  &$ (\frac{h}{4} , \frac{w}{4} , 256)\rightarrow (\frac{h}{4} , \frac{w}{4} , 256)$   &AdaResBlock-(N256, K3, S1, P1), AdaLIN, ReLU   \\ \cline{2-3} 
                  &$ (\frac{h}{4} , \frac{w}{4} , 256)\rightarrow (\frac{h}{4} , \frac{w}{4} , 256)$   & AdaResBlock-(N256, K3, S1, P1), AdaLIN, ReLU  \\ \cline{2-3} 
                  & $ (\frac{h}{4} , \frac{w}{4} , 256)\rightarrow (\frac{h}{4} , \frac{w}{4} , 256)$  &AdaResBlock-(N256, K3, S1, P1), AdaLIN, ReLU   \\ \hline
\multirow{3}{*}{Up-sampling } & $ (\frac{h}{4} , \frac{w}{4} , 256)\rightarrow (\frac{h}{2} , \frac{w}{2} , 128)$   &Sub-pixel-CONV-(N128, K3, S1, P1), LIN, ReLU   \\ \cline{2-3} 
                  & $ (\frac{h}{2} , \frac{w}{2} , 128)\rightarrow (h , w , 64)$  &Sub-pixel-CONV-(N64, K3, S1, P1), LIN, ReLU   \\ \cline{2-3} 
                  & $ (h , w , 64)\rightarrow (h , w , 3)$ &CONV-(N3, K7, S1, P3), Tanh   \\ \hline
\end{tabular}
\end{center}
\end{table*}

\begin{table*}[htp]
\caption{{\bf RA Analysis.} $a(x)=\gamma w E_x(x) + E_x(x)$, where the trainable parameter $\gamma$ determines the trade-off between the attended features and the original ones. When $\gamma=0$, it returns to $E_x(x)$ indicating no attention is used, and otherwise, the attention is activated.}
\begin{center}
\label{tab:gamma}
\begin{tabular}{|l|c|c|c|c|}
\hline
Object &  dog & winter &  photo & zebra \\
\hline
$\gamma$ &-0.2492	&0.2588	&-0.0006	&-0.2699 \\
\hline\hline
Object &  cat & summer &  vangogh & horse \\
\hline
$\gamma$  &0.3023	&0.0006	&0.3301	&0.2723 \\
\hline
\end{tabular}
\end{center}
\end{table*}

\subsection{Additional results}
\subsubsection{Discussing $\gamma$}
\setlength{\parindent}{1em}
As for Residual Attention (RA) module, the parameter $\gamma$ is task-specific (as illustrated in table~\ref{tab:gamma}). Regarding tasks like photo $\rightarrow$ vangogh and summer$\rightarrow$ winter, $\gamma$ is close to 0 indicating more attention is paid to global features, which is reasonable as translating the whole content of the images in these tasks is more necessary than focusing on local details.

\begin{table*}[htp]
\caption{{\bf Multi-Scale Analysis.} For both FID and KID, lower is better. Results of methods are all in 100K iterations of discriminator. }
\begin{center}

  \label{tab:Multi-Scale Analysis}
    \begin{tabular}{|l|c|c|c|c|c|c|c|c|}
    \hline
    \multirow{2}{*}{\diagbox{Method}{Dataset}}& 
    \multicolumn{2}{c|}{dog2cat}&\multicolumn{2}{c|}{ winter2summer } &\multicolumn{2}{c|}{ photo2vangogh} &\multicolumn{2}{c|}{zebra2horse}\cr\cline{2-9}
      & FID & KID $\times$ 100 & FID & KID $\times$ 100 & FID & KID $\times$ 100 & FID & KID $\times$ 100 \cr
    \hline \hline
      $C^0_{x}$  &216.03	&18.57	&81.12	&1.50	&135.17	&3.92	&215.79	&12.79 \cr\cline{1-9}
      $C^0_{x}$, $C^1_{x}$   &203.56	&15.27	&77.52	&1.14	&121.47	&2.86	&193.11	&10.37 \cr\cline{1-9}
      $C^0_{x}$, $C^1_{x}$, $C^2_{x}$ &48.79	&1.58	&76.44	&1.22	&122.27	&3.71	&149.48	&4.29 \cr\cline{1-9}
        $C^1_{x}$, $C^1_{x}$ &45.46	&0.85	&77.50	&1.17	&131.38	&5.38	&147.24	&3.92 \cr\cline{1-9}
        $C^0_{x}$, $C^2_{x}$ &54.31	&2.20	&88.02	&2.45	&130.73	&4.87	&154.13	&5.43 \cr\cline{1-9}
    \hline \hline 
    \multirow{2}{*}{\diagbox{Method}{Dataset}}& 
    \multicolumn{2}{c|}{cat2dog}&\multicolumn{2}{c|}{ summer2winter } &\multicolumn{2}{c|}{ vangogh2photo} &\multicolumn{2}{c|}{horse2zebra}\cr\cline{2-9}
      & FID & KID $\times$ 100 & FID & KID $\times$ 100 & FID & KID $\times$ 100 & FID & KID $\times$ 100 \cr
    \hline \hline
    $C^0_{x}$ 	&231.24	&22.12	&76.88	&0.63	&155.50	&7.40	&168.57	&10.74  \cr\cline{1-9}
    $C^0_{x}$, $C^1_{x}$	&238.62	&21.41	&77.10	&0.67	&132.08	&4.67	&104.46	&4.60  \cr\cline{1-9}
    $C^0_{x}$, $C^1_{x}$, $C^2_{x}$	&44.67	&1.20	&76.03	&0.67	&112.00	&2.79	&65.93	&2.09 \cr\cline{1-9}
        $C^1_{x}$, $C^1_{x}$ &53.94 	&1.95 	&79.91 	&1.11 	&128.47 	&4.87 	&90.00 	&3.77  \cr\cline{1-9}
        $C^0_{x}$, $C^2_{x}$ &65.99 	&2.62 	&96.26 	&2.08 	&123.05 	&4.32 	&80.50 	&2.85  \cr\cline{1-9}
    \end{tabular}
\end{center}
\end{table*}

\subsubsection{More analysis on the multi-scale discriminator.}

Table~\ref{tab:Multi-Scale Analysis} evaluates the impact of $(C^0_{x} , C^1_{x}, C^2_{x})$ on various datasets. For the cat $\leftrightarrow$ dog task, global characteristics of the semantic of objects is of much importance. 
For the colorization and stylization task(\emph{e.g.} summer $\leftrightarrow$ winter, photo $\leftrightarrow$ vangogh ), preserving middle and local scale still delivers promising performance. Specifically, if removing the local scale, FID increases significantly from 66 to 90 on \emph{horse$\rightarrow$zebra}; and from 76/76 to 88/96 on \emph{summer$\leftrightarrow$winter} if leaving out the medium scale. It implies all three scales are generally necessary. 

\begin{figure*}[htp]
\begin{center}
\includegraphics[width=1.0\linewidth]{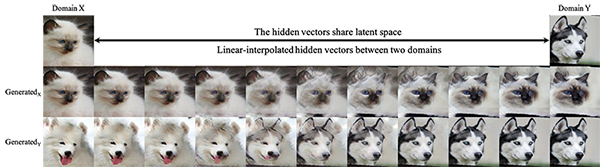}
\end{center}
  \caption{{\bf Translation results with linear-interpolated hidden vectors between two domains.}  ${\rm Generated_X}$: images of Domain X generated from the hidden vectors; ${\rm Generated_Y}$: images of Domain Y generated from the hidden vectors. Results show that the hidden vectors share latent space since it successfully generates reasonable image from linear-interpolated hidden vectors between two domains.}
\label{fig:share}
\end{figure*}

\begin{figure*}[htp]
\begin{center}
\includegraphics[width=0.8\linewidth]{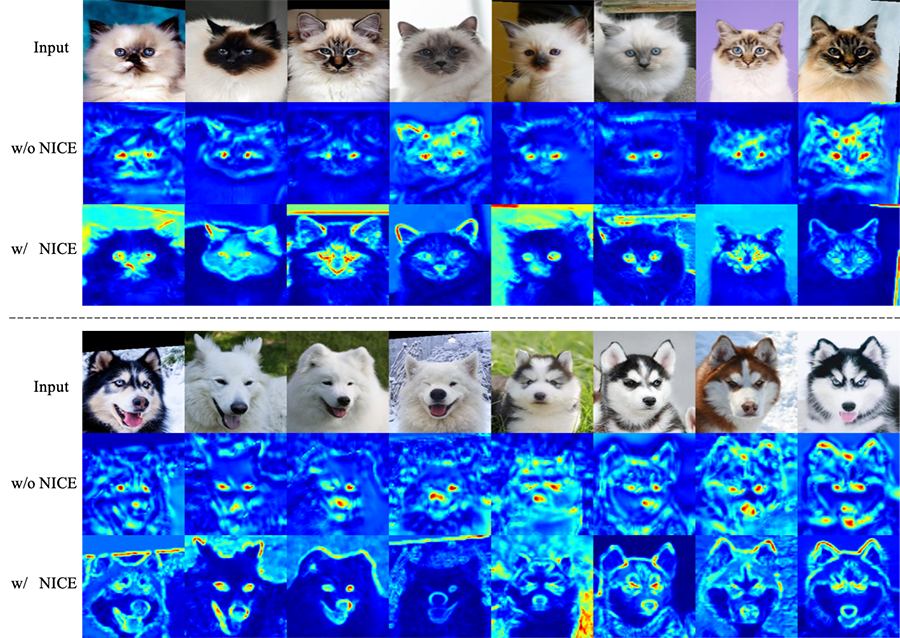}
\end{center}
  \caption{The heat-map visualizations of the hidden vectors. }
\label{fig:heatmaps}
\end{figure*}

\subsubsection{More visualizations of hidden vectors.}

 The training process is proceeded in terms of three kinds of losses: adversarial loss, identity reconstruction loss, and cycle-consistency loss. The adversarial loss is to pursue domain transfer, while both reconstruction loss and cycle-consistency loss are for tackling the non-identifiability issue. As shown in Figure~\ref{fig:share}, our method enables meaningful hidden interpolations since the shared-latent space assumption are enforced by NICE framework and three kinds of losses in our training.
 

Figure~\ref{fig:heatmaps} visualizes more heat-maps of the hidden vectors. Generally, the heat-maps by the model with NICE show more concise and distinguishable semantics encoding than that without NICE (namely an independent encoder is used). It shows using NICE captures the texture and local parts of the object more clearly, exhibiting the superiority of NICE-GAN.

\subsubsection{Additional comparisons with state of the arts}

Due to the lack of standard protocol so far, our experiments use released codes to train all baselines over the same iterations for fair comparison. 
Table~\ref{tab:Ad_FID_KID} shows additional comparisons with state of the arts in 200K-th iterations. Still, NICE-GAN (trained for more iterations) generally performs superiorly.

\begin{table*}[htp]
\caption{The FID and the KID $\times$100 for different algorithms. Lower is better. All of the methods are trained to the 200K-th iterations. }
\begin{center}

  \label{tab:Ad_FID_KID}
    \begin{tabular}{|l|c|c|c|c|c|c|c|c|}
    \hline
    \multirow{2}{*}{\diagbox{Method}{Dataset}}& 
    \multicolumn{2}{c|}{dog $\rightarrow$ cat}&\multicolumn{2}{c|}{winter $\rightarrow$ summer} &\multicolumn{2}{c|}{photo $\rightarrow$ vangogh} &\multicolumn{2}{c|}{zebra $\rightarrow$ horse}\cr\cline{2-9}
      & FID & KID $\times$ 100 & FID & KID $\times$ 100 & FID & KID $\times$ 100 & FID & KID $\times$ 100 \cr
    \hline \hline
      NICE-GAN   & {\bf 42.22}	& {\bf 0.73}	& 77.51	&1.37	&126.29	& 4.35	&{\bf 138.77}	 &{\bf3.26} \cr\cline{1-9}
      U-GAT-IT-light &63.85	&2.08	&{\bf 72.58}	&1.99	&120.92	&3.68	&150.34	&3.64 \cr\cline{1-9}
      CycleGAN &93.72	&3.46	&77.01	&{\bf 1.07}	&{\bf 115.74}	&{\bf 2.90}	&140.65	&3.64 \cr\cline{1-9}
      UNIT  &53.18	&1.36	&95.76	&4.59	&135.37	&5.03	&174.65	&6.36 \cr\cline{1-9}
        MUNIT   &48.52	&1.21	&99.14	&4.36	&132.22	&4.75	&190.06	&6.32	\cr\cline{1-9}
        DRIT   &63.13	&2.75	&83.30	&2.03	&126.11	&4.28	&164.92	&6.78\cr\cline{1-9}
    \hline \hline 
    \multirow{2}{*}{\diagbox{Method}{Dataset}}& 
    \multicolumn{2}{c|}{cat $\rightarrow$ dog}&\multicolumn{2}{c|}{summer $\rightarrow$ winter } &\multicolumn{2}{c|}{vangogh $\rightarrow$ photo} &\multicolumn{2}{c|}{horse $\rightarrow$ zebra}\cr\cline{2-9}
      & FID & KID $\times$ 100 & FID & KID $\times$ 100 & FID & KID $\times$ 100 & FID & KID $\times$ 100 \cr
    \hline \hline
    NICE-GAN	&{\bf 34.71}	&{\bf 0.61}	&78.87	&{\bf 0.78}	&{\bf 107.53}	&{\bf 2.99}	&75.64	&1.77  \cr\cline{1-9}
    U-GAT-IT-light	&69.43	&2.48	&84.16	&1.16	&110.03	&3.54	&85.66	&2.78 \cr\cline{1-9}
    CycleGAN	&103.95	&5.41	&{\bf 78.39}	&0.82	&117.88	&3.08	&{\bf 68.11}	&{\bf 1.52} \cr\cline{1-9}
    UNIT	&42.32	&0.90	&111.14	&5.34	&125.85	&5.97	&118.98	&6.34 \cr\cline{1-9}
    MUNIT	&45.17	&1.14	&110.91	&4.90	&131.25	&6.01	&104.72	&5.26 \cr\cline{1-9}
    DRIT	&53.19	&1.73	&81.64	&1.10	&111.46	&3.76	&92.26	&4.58 \cr\cline{1-9}
    \end{tabular}
\end{center}
\end{table*}

\subsubsection{More visualizations of translated images.}

\begin{figure*}[htp]
\begin{center}
\includegraphics[width=0.7\linewidth]{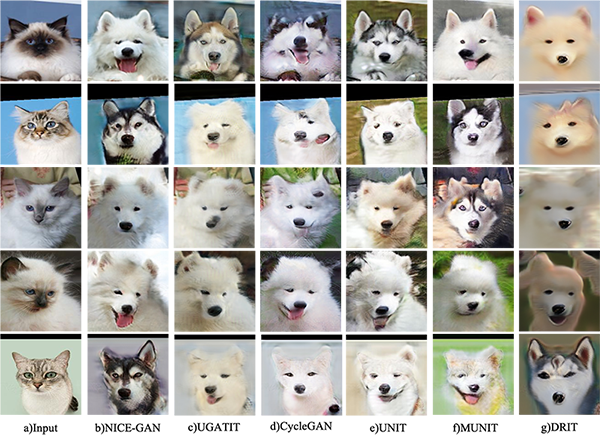}
\end{center}
  \caption{{\bf Examples of cat $\rightarrow $ dog translation images.} As is shown in these examples, images generated by NICE-GAN, UNIT and MUNIT have better quality.}
\label{fig:cat2dog}
\end{figure*}

\begin{figure*}[htp]
\begin{center}
\includegraphics[width=0.7\linewidth]{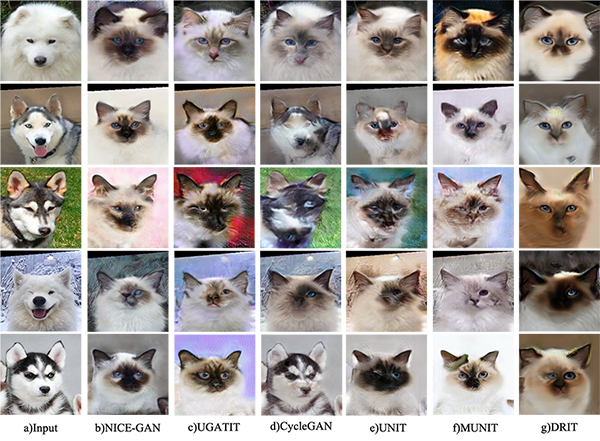}
\end{center}
  \caption{{\bf Examples of dog $\rightarrow $ cat translation images.} Most images are optimistic except those generated by CycleGAN and DRIT.}
\label{fig:dog2cat}
\end{figure*}

\begin{figure*}[htp]
\begin{center}
\includegraphics[width=0.7\linewidth]{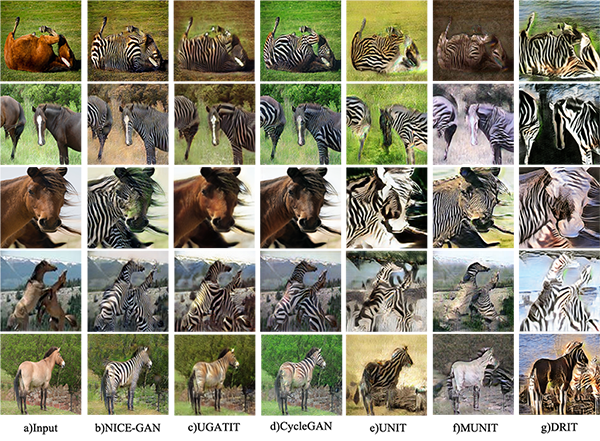}
\end{center}
  \caption{{\bf Examples of horse $\rightarrow $ zebra translation images.} The translation images shows that NICE-GAN has better ability in adding textures except for subtle color differences during the translation process. }
\label{fig:horse2zebra}
\end{figure*}

\begin{figure*}[htp]
\begin{center}
\includegraphics[width=0.7\linewidth]{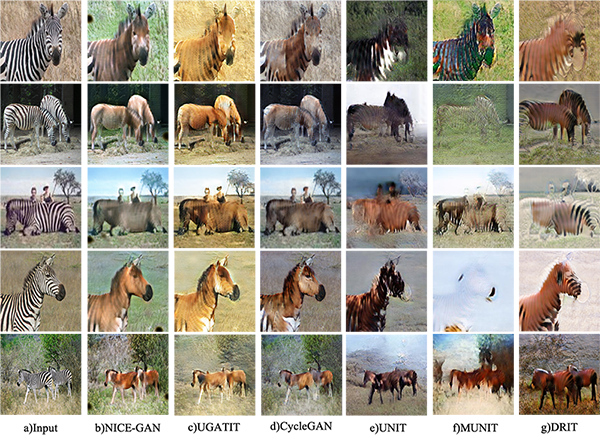}
\end{center}
  \caption{{\bf Examples of zebra $\rightarrow $ horse translation images.} As is shown in the examples, images generated by U-GAT-IT gain the best results. The disadvantage of NICE-GAN still lies in subtle color differences. }
\label{fig:zebra2horse}
\end{figure*}

\begin{figure*}[htp]
\begin{center}
\includegraphics[width=0.7\linewidth]{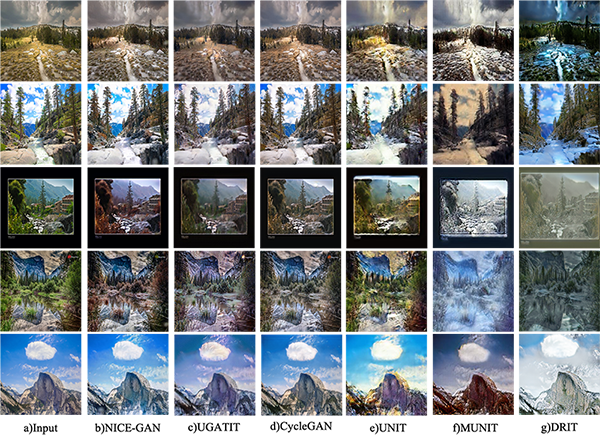}
\end{center}
  \caption{{\bf Examples of summer $\rightarrow $ winter translation images.} Images generated by different methods gain relatively ideal and realistic results.}
\label{fig:summer2winter}
\end{figure*}

\begin{figure*}[htp]
\begin{center}
\includegraphics[width=0.7\linewidth]{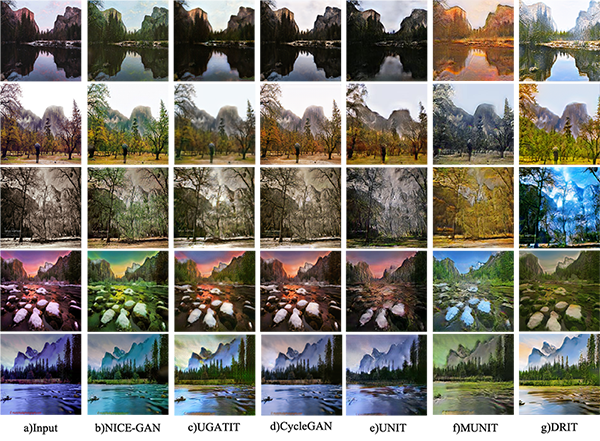}
\end{center}
  \caption{{\bf Examples of winter$\rightarrow $ summer translation images.} Images generated by different methods look optimistic except for images generated by CycleGAN and UNIT. 
 }
\label{fig:winter2summer}
\end{figure*}

\begin{figure*}[htp]
\begin{center}
\includegraphics[width=0.7\linewidth]{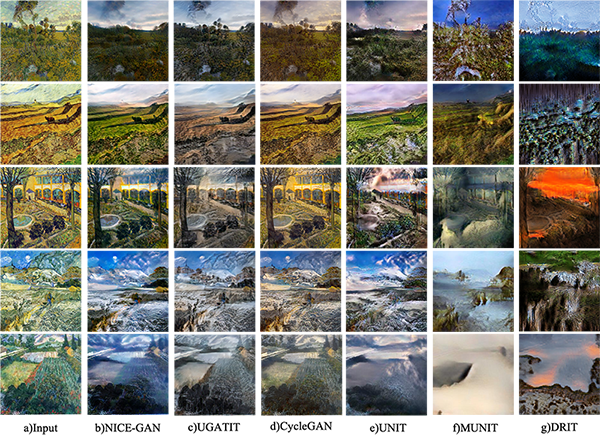}
\end{center}
  \caption{{\bf Examples of vangogh $ \rightarrow $ photo translation images.} The translation of vangogh $ \rightarrow $ photo is a difficult task, most methods could barely finish the task.
 }
\label{fig:van2pho}
\end{figure*}

\begin{figure*}[htp]
\begin{center}
\includegraphics[width=0.7\linewidth]{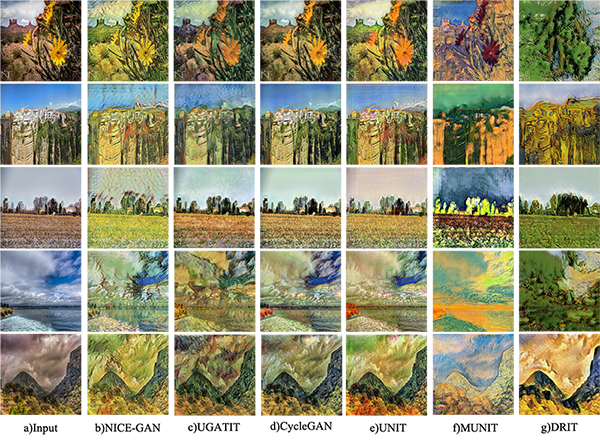}
\end{center}
  \caption{{\bf Examples of photo $\rightarrow $ vangogh translation images.} Images generated by different methods gain relatively ideal results except for DRIT.
 }
\label{fig:pho2van}
\end{figure*}

In addition to the results presented in the paper, we show more generated images for the four datasets in Figure~\ref{fig:cat2dog}, \ref{fig:dog2cat}, \ref{fig:horse2zebra}, \ref{fig:zebra2horse}, \ref{fig:summer2winter}, \ref{fig:winter2summer}, \ref{fig:van2pho} and \ref{fig:pho2van}.

\end{appendix}

\end{document}